\documentclass[10pt]{article} % For LaTeX2e
\usepackage[preprint]{tmlr}

\usepackage{hyperref}
\usepackage{url}
\usepackage{booktabs}
\usepackage{times}
\usepackage{soul}
\usepackage{url}
\usepackage{mathtools}
\usepackage[mathscr]{euscript}
\usepackage{amsmath}
\usepackage{amsthm}
\usepackage{booktabs}
\usepackage{algorithm}
\usepackage{algorithmic}
\usepackage{graphicx}
\usepackage{bbm}
\usepackage{multirow}
\usepackage{amsfonts}

\title{Graph Harmony: Denoising and Nuclear-Norm Wasserstein Adaptation for Enhanced Domain Transfer in Graph-Structured Data}

\author{\name Mengxi Wu \email mengxiwu@usc.edu \\
      \addr USC Computer Science Department \\ 
      \name Mohammad Rostami \email rostamim@usu.edu \\
      USC Computer Science Department 
      }

\def\month{MM}  % Insert correct month for camera-ready version
\def\year{YYYY} % Insert correct year for camera-ready version
\def\openreview{\url{https://openreview.net/forum?id=XXXX}} % Insert correct link to OpenReview for camera-ready version

\begin{document}

\maketitle

\begin{abstract}
% Domain adaptation is a fuDNANmental approach for machine learning tasks under data-scarce schemes. However, graph domain adaptation methods have not been widely explored yet. Current graph domain adaptation techniques mainly combine adversarial methods and graph neural networks. Instead of using a domain discriminator in this work, we align the target domain distribution with an annotated source domain distribution in a shared embedding space. Specifically, we propose minimizing both the Sliced Wasserstein Distance and the Maximum Mean Discrepancy between source and target domain features class-wisely. We also develop a new and reliable pseudo-label generation technique to achieve class-wise feature alignment. A data augmentation technique for graph-structured data is adopted to enhance performance further. Extensive experiments on the Ego-network and the IMDB$\&$Reddit datasets show that our method is competitive and could outperform current state-of-the-art methods.

%Deep learning models present significant challenges due to the high costs associated with collecting and annotating large training data sets. 

Graph-structured data can be found in numerous domains, yet the scarcity of labeled instances hinders its effective utilization of deep learning in many scenarios. Traditional unsupervised domain adaptation (UDA) strategies for graphs primarily hinge on adversarial learning and pseudo-labeling. These approaches fail to effectively leverage graph discriminative features, leading to class mismatching and unreliable label quality. To navigate these obstacles, we develop the Denoising and Nuclear-Norm Wasserstein Adaptation Network (DNAN). DNAN employs the Nuclear-norm Wasserstein discrepancy (NWD), which can simultaneously achieve domain alignment and class distinguishment. DANA also integrates a denoising mechanism via a variational graph autoencoder that mitigates data noise. This denoising mechanism helps capture essential features of both source and target domains, improving the robustness of the domain adaptation process. Our comprehensive experiments demonstrate that DNAN outperforms state-of-the-art methods on standard UDA benchmarks for graph classification.% validating its effectiveness in domain adaptability. 

\end{abstract}

\section{Introduction}
While deep learning has made substantial progress in handling graph-structured data, it shares a substantial drawback with other methods in the same category — a heavy reliance on labeled data. This requirement presents a significant obstacle in real-world applications, where the gathering and annotating of graph-structured data come with a steep price tag, both in terms of time and resources. Sometimes, obtaining the labels is impossible because of privacy concerns. Sometimes, obtaining detailed labels for graph-structured data, e.g.,  chemical molecules, is a considerable challenge, e.g. because chemical molecules are incredibly complex, comprising a large number of atoms connected in various ways through different kinds of bonds. Understanding relations in a graph precisely and figuring out the individual attributes of graph nodes is not easy to achieve. Collecting annotated graph-structured data like social networks is also challenging due to the need to protect personal and sensitive information and the continual changes in network relationships. The label scarcity makes it difficult to derive meaningful insights and hinders the development of strategies and solutions based on deep learning. Therefore, it is highly desirable to relax the need for extensive graph-structured data annotation to replicate the success of deep learning in applications with graph-structured data. 

To navigate the challenge of label scarcity, Unsupervised Domain Adaptation (UDA) ~\cite{ganin2015unsupervised} has emerged as a promising frontier, with the aim of leveraging labeled data from a related source domain to inform an unlabeled target domain ~\cite{ramponi2020neural}. The principle of UDA is to align the data distributions between the two domains within a common embedding space, enabling a classifier trained on the source domain to perform competently on the target domain~\cite{rostami2023domain}. While UDA has been extensively applied to array-structured data ~\cite{long2016unsupervised,kang2019contrastive,rostami2019deep}, its translation to graph-structured data remains under-explored. Pioneering methods, such as DANE ~\cite{zhang2019dane}, integrate generative adversarial networks (GANs) with graph convolutional networks (GCNs) to align the domains. Others, like the approach by Wu et al. ~\cite{wu2020unsupervised}, introduce attention mechanisms to reconcile global and local consistencies, again employing GANs for cross-domain node embedding extraction. These GAN-based methods have the drawback of class mismatching, lacking clear separability between features from different classes, as they align target and source domain features irrespective of their classes.

Our work benefits from the Denoising VGAEs and Nuclear-Norm Wasserstein Adaptation Network (DNAN) and innovates on the existing UDA landscape by specifically addressing the class mismatchment issue for existing graph UDA methods. By leveraging the DNAN, unlike the previous GAN-based method, DNAN performs a refined, class-specific alignment of source and target domain distributions within a shared embedding space, preserving the distinct separability of features across classes.  Moreover, we incorporate a denoising mechanism through a variational graph autoencoder (VGAE). This inclusion of denoising is crucial as it alleviates the noise intrinsic to graph data and enhances feature representation for transferability. By using these two components, DNAN performs competitively and achieves SOTA performance on major UDA benchmarks for graph classification.

\section{Related Work}
%Our work is at the intersection of  the representation of graph-structured data using suitable neural networks and UDA.

\textbf{Unsupervised Domain Adaptation}
A foundational approach within UDA is to reduce the discrepancy between the source and target domain distributions using adversarial learning. A seminal method in this space, the Domain Adversarial Neural Network (DANN) \cite{ganin2015unsupervised}, employs an adversarial training framework to align domain representations by confusing a domain classifier in a shared embedding space. This strategy is adapted from generative adversarial networks (GANs) \cite{goodfellow2020generative}, tailored for domain adaptation purposes. Expanding on this adversarial methodology, the FGDA technique \cite{gao2021gradient} used a discriminator to discern the gradient distribution of features, thereby achieving better performance in mitigating domain discrepancy. Furthermore, DADA \cite{tang2020discriminative} proposed an innovative strategy by integrating the domain-specific classifier with the domain discriminator to align the joint distributions of two domains more effectively.

Although adversarial approaches are effective, they are complemented by statistical discrepancy measures like Maximum Mean Discrepancy (MMD), utilized in the Joint Distribution Optimal Transport (JDOT) model \cite{courty2017joint}. Although MMD effectively measures distributional divergence, it may not capture higher statistical moments, an area where the Wasserstein distance (WD) \cite{villani2008optimal} excels. WD has been leveraged for distribution alignment in UDA \cite{courty2017optimal,rostami2023overcoming, damodaran2018deepjdot}, with Redko et al. \cite{redko2017theoretical} providing theoretical underpinnings for model generalization on the target domain when employing WD. However, the practical application of WD is computationally intensive due to the absence of a closed-form solution. The Sliced Wasserstein Distance (SWD) \cite{rabin2011bwasserstein, bonneel2015sliced,rostami2021lifelong,rostami2022increasing} offers a computationally feasible alternative. Reconstruction-based objectives constitute another research direction, enforcing feature invariance across domains by reconstructing source domain data from target domain features, as in the work by Ghifary et al. \citeyear{ghifary2016deep}. Additionally, the application of   autoencoders  to UDA~\cite{rostami2021cognitively}, such as in the Variational Fair Autoencoder \cite{louizos2015variational}, showcases the capabilities of probabilistic generative models in domain-invariant feature learning. Our proposed method draws inspiration from the variational autoencoder's framework. Other notable approaches like ToAlign \cite{wei2021toalign}, SDAT\cite{rangwani2022closer}, and BIWA \cite{biwaa2023}, mark the recent advancement in UDA, surpassing previous models in performance. These three approaches are detailed in the experiment sections as our references for current state-of-the-art methods. However, extending these existing methods to graph-structured is often non-trivial.

 \textbf{Graph Representation Learning}
%The major obstacle in using deep learning for graph-structured data is representing graphs via vectors that encode graph structures. %If we map the edges and nodes of a graph into low-dimensional vectors such that   graph structures and geometrical properties are maintained, we can use the vectors as inputs to complex deep neural networks.
%A group of methods has tried to benefit from deep learning to learn descriptive embeddings for graph-structured data~\cite{wang2019attributed,pan2019learning}. In contrast, g
% Graph neural networks are generally extensions of well-known architectures for array-structured data to process graph-structured data directly. They convert a graph into a dense vector that can then be processed similarly to array-structured data. 
% Graph convolutional networks (GCN) \cite{kipf2016semi} is an extension of convolutional neural network (CNN), which integrates node features and graph topology based using the graph adjacency matrix.  The graph convolutional and global pooling layers compute the global embeddings for inputs. GCNs have been extremely successful in improving performance on a wide range of learning tasks for graph-structured data~\cite{zhang2019graph}.
% Graph attention networks (GATs)~\cite{velivckovic2017graph} are improvements over GCNs which benefit from an attention mechanism to learn the weights that aggregate features from the local neighborhood of a node.
Graph representation learning (GRL) has emerged as an important approach in machine learning, tasked with distilling complex graph-structured data into a tractable, low-dimensional vector space to enable using architectures developed for array-structured data. Previously, seminal spectral methods laid the foundation, leveraging graph Laplacians to capture topological structures of graphs despite the limitations in scalability for larger graphs \cite{belkin2003laplacian, chung1997spectral}. The field then evolved with algorithms such as DeepWalk and Node2Vec, which utilized random walks to encode local neighborhood structures into node embeddings, balancing the preservation of local and global graph characteristics \cite{perozzi2014deepwalk,grover2016node2vec}. The introduction of Graph Neural Networks (GNNs) marked a significant advancement in GRL. GNNs, specifically Graph Convolutional Networks (GCNs), offer a means to generalize neural network approaches to graph data, integrating neighborhood information into node embeddings \cite{kipf2016semi}. This was further refined by GraphSAGE, which scaled GNNs by learning a function to sample and aggregate local neighborhood features \cite{hamilton2017representation,hamilton2017inductive}. Moreover, Graph Attention Networks (GATs) introduced an attention mechanism, enabling the model to adaptively prioritize information from different parts of a node's neighborhood, thus enhancing the expressiveness of the embeddings \cite{velickovic2017graph}. These advances, along with the development of Graph Autoencoders like VGAEs that focused on graph reconstruction from embeddings, have broadened the applications of GRL and continue to shape its trajectory \cite{kipf2016variational}. For a fair comparison, when we compare with methods originally not proposed for graph domain adaptation, we replace their feature extraction backbones with GATs.

% A major approach to address UDA for array-structured data is to select a suitable probability metric   and then align the distributions by minimizing the distance between the source and the target distributions~\cite{courty2017optimal}.   The Maximum Mean Discrepancy (MMD)  is of the earliest metrics   used for this purpose~\cite{gretton2009covariate}. MMD  measures  the distances between two distributions   as the distance between their statistical mean. 
% While MMD misses matching the higher statistical moments between two distributions, the Wasserstein distance (WD)~\cite{villani2008optimal} has been used for distribution alignment in UDA~\cite{courty2017optimal,damodaran2018deepjdot} to match the higher statistical moments. 
% Redko et al.~\cite{redko2017theoretical} provided theoretical guarantees on model generalization on the target domain when WD is used.  An obstacle to using   WD is that it does not have a closed-form solution, and its computation is   computationally demanding. The Sliced Wasserstein Distance (SWD)~\cite{rabin2011bwasserstein,bonneel2015sliced} is an alternative metric to relax the need for solving this optimization problem. % SWD is computed based on the idea of slicing  high-dimensional distributions using  1D projections, computing the Wasserstein distance between the 1D slices, and averaging results along several random slices. 
%Since WD has a closed-form solution, it can be computed efficiently. 
% We benefit from both MMD and SWD to help align both high-order and low-order statistical moments of the two distributions.

\section{Problem Description}
We operate under the assumption that there is a source domain with labeled data and a target domain with exclusively unlabeled data. In both domains, each input instance comprises a graph-structured data sample. Our primary objective is to develop a predictive model for the target domain by transferring knowledge from the source domain.% capable of performing classification   within the target domain by leveraging knowledge transfer from the source domain. %This approach aims to facilitate improved performance and accuracy in the target domain, despite the absence of labeled data.

\textbf{Graph Classification} 
 We focus on a graph classification task, where a graph sample can be represented as $G= (X, A)$. $X \in \mathbb{R}^{n\times K}$, where $n$ represents the number of nodes in $G$ and $K$ represents the dimension of the features for each node.  Specifically, $x_i \in X$ corresponds to the feature associated with a node $v_i$. Let $A \in \mathbb{R}^{n\times n}$ denote the adjacency matrix. The matrix $A$ encapsulates the topological structure of $G$.  Each graph is associated with a class, and we use $y$ to denote the ground truth label of graph sample $G$. The goal is to train a model capable of classifying graphs effectively and accurately.

\textbf{Source Domain Dataset} We consider a fully labeled source domain dataset as $(D_s, Y_s) = (\{G_s^k\}, \{y_s^k\})$,  where $G_s^k$ is the $k^{th}$ graph sample in $D_s$ and $y_s^k$ is the ground truth label of $G_s^k$.

\textbf{Target Domain Dataset} We consider that only an unlabeled target domain dataset $D_t = \{G_t^k\}$ is accessible, where $G_t^k$ is the $k^{th}$ graph sample in the target dataset $D_t$.
The target domain is our problem of interest.

\textbf{UDA for Graph Classification} 
Given an unlabeled target domain dataset $D_t$ and a fully labeled source domain dataset $D_s$, our goal is to develop an effective approach to transfer knowledge from the source domain graph samples to the target domain graph samples, enabling accurate classification in the target domain.% domain despite the lack of labeled data.

\section{Proposed Method}

\begin{figure*}[!ht]
\centering
\includegraphics[width=16.6cm]{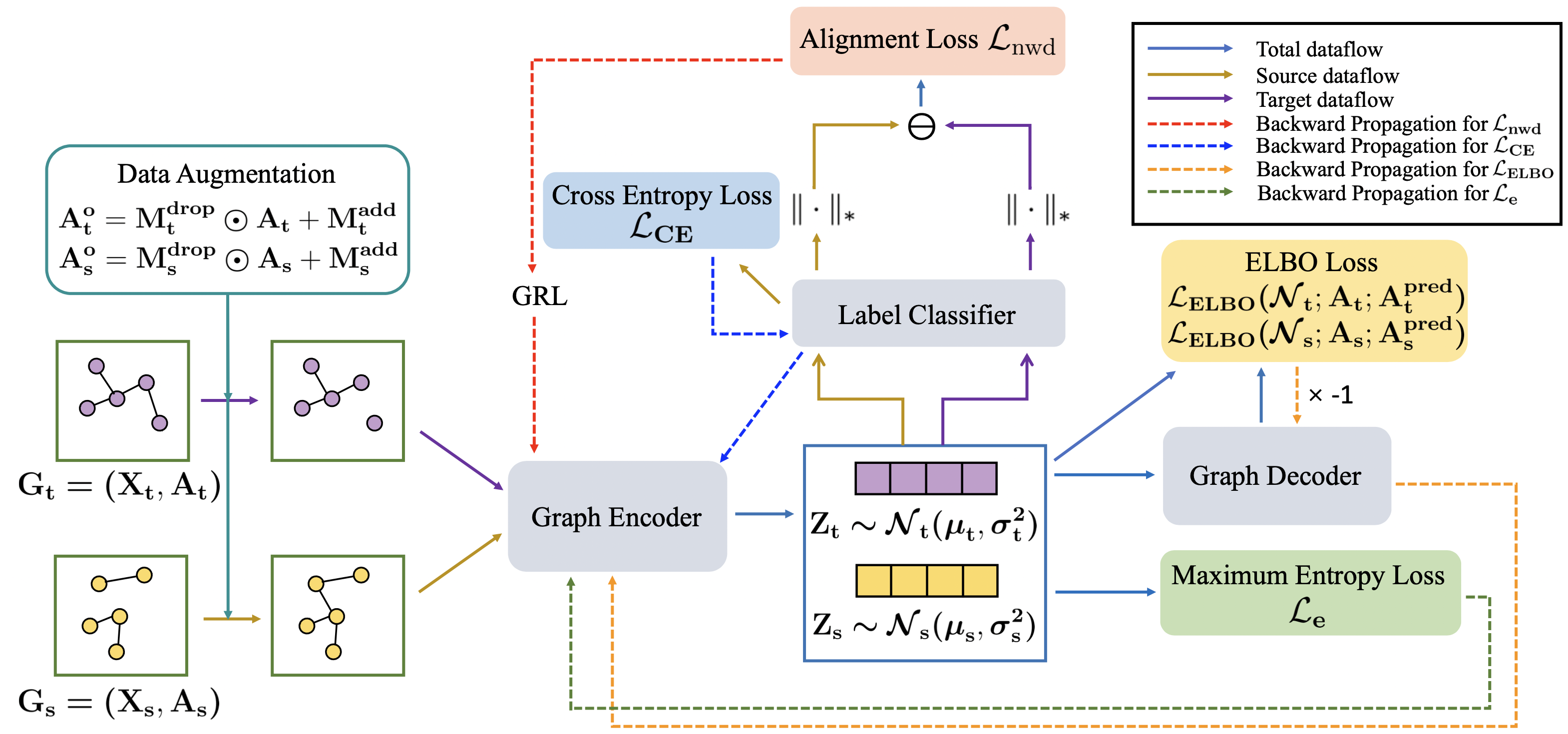}
\caption{The block-diagram visualization of DNAN: We first add noise to the graph structures of the source and target domains by applying data augmentation to their adjacency matrices \(A_s\) and \(A_t\), using masks \(M_t^{\text{drop}}\), \(M_s^{\text{drop}}\), \(M_s^{\text{add}}\), and \(M_t^{\text{add}}\). Then, the graph encoder of a VGAE produces the latent variable \(Z_s\) and \(Z_t\) from node features \(X_s, X_t\) and augmented adjacency matrices \(A_s^o, A_t^o\). We train a label classifier using a cross-entropy loss \(\mathcal{L}_{\text{CE}}\) between the output of the label classifier from $Z_s$ and the ground-truth label. To align the latent variables of both domains, we compute a Nuclear-norm Wasserstein discrepancy (NWD) alignment loss using \(Z_s\), \(Z_t\), and the label classifier. The VGAE also includes a graph decoder that reconstructs the original adjacency matrices \(A_s, A_t\) from \(Z_s, Z_t\). Then, the Evidence Lower Bound (ELBO) loss is computed based on the outputs of the graph encoder and original and reconstructed adjacency matrices. Lastly, the model applies maximum Entropy regularization \(L_e\) to the latent variables $Z_t, Z_s$.} % in accordance with Eq.\eqref{eq2}.} 
\label{fig1}
\end{figure*}

Figure \ref{fig1} visualizes a high-level description of our proposed pipeline. Our algorithm benefits from a denoising mechanism via variational graph autoencoder and the Nuclear-norm Wasserstein discrepancy for distribution alignment. In a nutshell, our solution is to effectively embed the graph structures from both domains into a shared feature space and then align the distributions of both domains in the shared feature space by minimizing a suitable distribution metric.

\subsection{Latent Variables Reconstruction}
Our model adopts encoding and decoding structures from the Variational Graph Auto-Encoders (VGAE) \cite{kipf2016variational}, an unsupervised learning model for graph-structured data based on Variational Auto-Encoder (VAE) concepts. Given a graph sample $G = (X, A)$ with $n$ nodes, the encoder in VGAE generates a corresponding latent variable $Z$.  \( q_\phi(Z|A, X) \) is used to denote the feature encoder, characterized by the parameter \( \phi \). \(q_\phi\) aims to approximate the real posterior distribution \( p(Z|A, X) \). The decoder of a standard VGAE is represented as \( P_\theta(A|Z) \), defined by parameters \( \theta \). The prior distribution is denoted by \( P(Z) \). The prior distribution is assumed to be a normal distribution, specifically \( P(Z) \sim \mathcal{N}(0,I) \). The variational lower bound of the marginal likelihood for standard VGAE is given as:
\begin{equation}
\mathcal{L}_{\text{ELBO}}(\phi, \theta) = E_{q_\phi(Z|A^o,X)}[P_\theta(A^o|Z)] - D_{KL}(q_\phi(Z|A^o,X)||P(Z)) 
\end{equation}
where \( D_{KL}(q(\cdot)||P(\cdot)) \) represents the Kullback-Leibler divergence between distributions \( q(\cdot) \) and \( P(\cdot)\). The standard VGAE is trained through maximizing the \(\mathcal{L}_{\text{ELBO}}\). Similar to the training process of VGAE, however, we have some variations. Given a graph sample $G=(X,A)$, we train the VGAE on both $G=(X,A)$ and $G_o(X,A^o)$, where $A_o$ denoted the augmented adjacency matrix. We add noise to the adjacency matrices to implement data augmentation. Specifically, we benefit from a random manipulation-based approach  \cite{cai2021graph}. %We randomly drop or add edges to every graph sample to generate the augmented version. %In other words, we introduce noise to the original graph. 
To this end, edges are dropped and added randomly by modifying the values in the adjacency matrix $A$ of the original graph. The modified adjacency matrix $A^o$ is constructed as follows:
\begin{equation}
\begin{split}
A^o = M^{\text{drop}} \odot A + M^{\text{add}}, \hspace{5mm}
m_{ij}^{\text{add}} \sim \text{Bernoulli}(p^\text{{add}} \cdot p^\text{{edge}}), \hspace{5mm}
m_{ij}^{\text{drop}} \sim \text{Bernoulli}(p^\text{{drop}})
\end{split}
\end{equation}
where $p^{\text{{add}}}$, $p^{\text{{edge}}}$, and $p^{\text{{drop}}}$ denote the edge addition rate, the sparsity of the adjacency matrix $A$, and the edge dropping rate. $\odot$ represents the element-wise multiplication between two matrices. $M^{\text{drop}}$ and $M^{\text{add}}$ represent mask matrices with the same dimensions as $A$. For each element $m_{ij}^{\text{add}} \in M^{\text{add}}$ or $m_{ij}^{\text{drop}} \in M^{\text{drop}}$,  we sample its value from a Bernoulli distribution.

In a standard Variational Graph Autoencoder setup, if the input is an augmented adjacency matrix $A^o$, then the VGAE reconstructs $A^o$. However, in our approach, we aim to reconstruct the original adjacency matrix before augmentation. In other words, we reconstruct $A$ given either $G=(X,A)$ or $G^o=(X,A^o)$. This way, we train the variational graph autoencoder to function as a sophisticated filter purifying the data. This purification is crucial for aligning the source and target domain distributions. The process can be likened to enhancing one's vision in dense fog: the feature encoder, through training, becomes proficient at distinguishing unwanted random information and can reliably identify and extract the core features fundamental to source and target domains. These core features, distilled by the denoising process, are then used to align the distributions of the source and target domains. By focusing on these essential features, we hypothesize that the distribution alignment can be based on the intrinsic similarities of the source and target domains rather than on noisy information not representative of the actual distribution. Following this idea, we modify the variational lower bound of the marginal likelihood to be:
\begin{equation}
\mathcal{L}_{\text{ELBO}} = E_{q_\phi(Z|A^o,X)}[P_\theta(A|Z)] - D_{KL}(q_\phi(Z|A^o,X)||P(Z)) 
\end{equation}
by replacing \(P_\theta(A^o|Z)\) with \(P_\theta(A|Z)\). We utilize Graph Attention Networks (GAT) as the encoder \( q_{\phi} \) of the VGAE. The encoding equations are given as follows:
\begin{equation}
\begin{split}
\mu &= \text{GAT}_\mu(A^o, X)\\
\log\sigma &= \text{GAT}_\sigma(A^o, X) \\
z_i  = \mu_i + &\varepsilon_i \cdot \sigma_i, \text{ } \varepsilon_i \sim \mathcal{N}(0,1)\\
q_{\phi}(z_{i}|A^o, X) &= \mathcal{N}(z_{i}|\mu_{i}, \text{diag}(\sigma_i^2))\\
q_{\phi}(Z|A^o, X) &= \prod_{i=1}^{n} q_{\phi}(z_{i}|A^o, X) 
\end{split}
\end{equation}
 The element \( z_{i} \) corresponds to the \( i^{th} \) row of \( Z \). This same row-wise correspondence applies to \( \mu_{i} \) and \( \log\sigma_{i} \) as well. Using the reparameterization trick, we transform the generated $\mu_i$ and $\sigma_i$ into latent variable \( z_i\).  After obtaining the latent variable $Z$, an inner product decoder \(P_{\theta}(A|Z) \) is applied to \( Z \) to reconstruct the adjacency matrix before data augmentation. This decoder translates each pair of node representations into a binary value, indicating whether an edge exists in the reconstructed adjacency matrix \( A' \). Specifically, we first use an MLP layer to enhance the expressive capacity of the latent variable $Z$ and then compute the dot product for each node representation pair as:
\begin{equation}
\begin{split}
 H &= \text{ReLU}((Z\cdot W_{0}) \cdot W_{1})\\
 p(A'_{ij} &= 1|h_{i}, h_{j}) = \sigma(h_{i}^Th_{j})\\
 p(A'|Z) &= \prod_{i=1}^{n} \prod_{j=1}^{n} p(A'_{ij}|h_{i}, h_{j})
 \end{split}
\end{equation}
where \( h_{i} \) represents the \( i^{th} \) row of \( H \) and \( A'_{ij} \) is an element of reconstructed adjacency matrix \( A' \). The decoder is described by parameter $\theta$ that includes the parameters \(\{ W_{0}, W_{1} \}\). To eliminate redundant information within the latent variables, an element-wise maximum entropy loss, denoted as \( L_e \), is applied to the latent variable $Z$. The regularization factors work to repel extraneous information from them, ensuring a cleaner reconstruction of the latent variable. The specifics of this regularization process are described below:
\begin{equation}
\begin{split}
\mathcal{L}_{\text{e}} &= \frac{1}{N_s + N_t} \sum_{G^k \in (D_s, D_t)} \text{ME}(Z^k)\\
\text{ME}(Z^k) &= \frac{1}{n_k \times F} \sum_{i=0}^{n_k} \sum_{j=0}^{F} \sigma(z_{ij}) \log \sigma(z_{ij})
\end{split}
\end{equation}
where $n_k$ is the number of nodes in the graph sample $G^k$ and \( F\) denotes the dimension of $G^k$'s latent variable $Z^k$.

% \subsection{Data Augmentation}
% We add noise to the adjacency matrices to implement data augmentation to alleviate the performance degradation caused by the domain gap on the target domain, i.e., we benefit from a random manipulation-based approach  \cite{cai2021graph} to generate data samples with noise. %We randomly drop or add edges to every graph sample to generate the augmented version. %In other words, we introduce noise to the original graph. 
% To this end, we drop and add edges randomly by modifying the values in the adjacency matrix $A$ of the original graph. We call the modified adjacency matrix $A^o$ and construct $A^o$ as:
% \begin{equation}
% \begin{split}
% A^o = M^{\text{drop}} \odot A + M^{\text{add}}, \hspace{5mm}
% m_{ij}^{\text{add}} \sim \text{Bernoulli}(p^\text{{add}} \cdot p^\text{{edge}}), \hspace{5mm}
% m_{ij}^{\text{drop}} \sim \text{Bernoulli}(p^\text{{drop}})
% \end{split}
% \end{equation}
% where $p^{\text{{add}}}$, $p^{\text{{edge}}}$, and $p^{\text{{drop}}}$ denote the edge addition rate, the sparsity of the adjacency matrix $A$, and the edge dropping rate. $\odot$ represents the element-wise multiplication between two matrices. $M^{\text{drop}}$ and $M^{\text{add}}$ represent mask matrices with the same dimensions as $A$. For each element $m_{ij}^{\text{add}} \in M^{\text{add}}$ or $m_{ij}^{\text{drop}} \in M^{\text{drop}}$,  we sample its value from a Bernoulli distribution. This process helps to improve model generalization during the initial training process.

\subsection{Distribution Alignment}
By introducing the new variational lower bound for the marginal likelihood, the VGAE encoder is better equipped to grasp the essential features of both domains. However, we still face a crucial challenge: addressing the performance degradation that occurs when a model trained on data from a source domain is applied to a target domain with a different data distribution. As mentioned in the previous sections, traditional approaches in unsupervised domain adaptation often use a domain discriminator that engages in a min-max game with a feature extractor to produce domain-invariant features. However, these methods primarily focus on confusing features at the domain level, which might negatively impact class-level information and lead to the mode collapse problem  \cite{kurmi2019looking, tang2020discriminative}. 
To address these challenges, our approach integrates the Nuclear-norm Wasserstein discrepancy (NWD) \cite{chen2022reusing} to effectively align the source and target domains' feature representations while maintaining class-level discrimination by considering it as a loss function to enforce domain alignment.  Our method utilizes a graph variational autoencoder described in the previous section and a classifier \(C\) with parameter \(\theta_c\). We construct \(C\) with two fully connected layers. The classifier \(C\) serves two purposes. As a classifier, it distinguishes between categories, and as a discriminator, it aligns features to close the gap. The empirical NWD loss is defined as:
\begin{equation}
 \mathcal{L}_{\text{nwd}} = \frac{1}{N_s^{\text{train}}}\sum_{k=1}^{N_s^{\text{tran}}} \|C(Z^k_s)\|_* - \frac{1}{N_t^{\text{train}}}\sum_{k=1}^{N_t^{\text{train}}}\|C(Z^k_t)\|_*
\end{equation}
 where $Z^k_s$ represents the latent variable for the graph sample $G^k_s$ and $Z^k_t$ represents the latent variable for the graph sample $G^k_t$. \(\|\cdot\|_*\) denotes the Nuclear norm. \(N_s^{\text{train}}\) is the number of training samples in the source dataset, and \(N_t^{\text{train}}\) is the number of training samples in the target dataset. To avoid complex alternating updates, we employ a Gradient Reverse Layer (GRL) \cite{ganin2016domain}, which allows for updating in a single backpropagation step without gradient penalties or weight clipping. The distribution alignment is achieved through a min-max game, optimized as:
\begin{equation}
 \min_{\phi} \max_{\theta_c} \mathcal{L}_{\text{nwd}}
\end{equation}

\begin{algorithm}[!ht]
    \caption{DNAN Method}
    \label{alg:algo1}
    \textbf{Input}: $(D_s, Y_s), D_t$\\
    \textbf{Parameters}: VGAE parameters $\{\phi\text{ (Encoder)}, \theta \text{ (Decoder)}\}$, Classifier parameter $\{\theta_c\}$\\
    \textbf{Output}: Trained Parameters $\phi, \theta, \theta_c$
    
    \begin{algorithmic}[1] %[1] enables line numbers
        \STATE Randomly sample a batch of $\{(G^k_s, y^k_s)\}$
        \STATE Randomly sample a batch of $\{G_k^t\}$
        \STATE Forward Propagation
        \STATE Update $\phi, \theta, \theta_c$ based on Equation (10)
        \STATE Augment $\{G^k_s\}, \{G^k_t\}$ based on Equation (2)
        \STATE Forward Propagation and set $\mathcal{L}_{\text{cls}}=0$
        \STATE Update $\phi, \theta, \theta_c$ based on Equation (10)
        \STATE \textbf{return}  $\phi, \theta, \theta_c$
    \end{algorithmic}
\end{algorithm}

\subsection{Algorithm Summary}
In addition to distribution alignment, to ensure accurate classification, we optimize the encoder in VGAE and the classifier \(C\) using a supervised classification loss \(\mathcal{L}_{\text{cls}}\) for the source domain:
\begin{equation}
 \mathcal{L}_{\text{cls}} = \frac{1}{N_s^{\text{train}}}\sum_{j=1}^{N_s^{\text{train}}}\mathcal{L}_{\text{CE}}(C(Z^k_s,y^k_s))
 \end{equation}
Then, by combining all the loss described in the previous sections, our total optimization object is formulated as follows:
\begin{equation}
\min_{\phi,\theta,\theta_c} \big \{ \mathcal{L}_{\text{cls}} -  \mathcal{L}_{\text{ELBO}}+   \lambda_e\mathcal{L}_{\text{e}}\big\} + \min_{\phi} \max_{\theta_c} \mathcal{L}_{\text{nwd}},
\end{equation}
where $\phi$ is the parameter of the feature encoder of the VAE, $\theta$ is the parameter of the decoder of the VAE, $\theta_c$ is the parameter of the classifier and $\lambda_e$ is a hyperparameter weigh the maximum entropy loss $\mathcal{L}_\text{e}$. It is worth noting that we balance the supervised classification loss and the NWD loss equally. In this case, our model learns transferable and discriminative features and has prediction accuracy and diversity in the target domain. The complete procedures of our UDA approach for graph-structured data are summarized in Algorithm ~\ref{alg:algo1}. 

\section{Experimental Validation}
We validate our algorithm using two graph classification benchmarks. Our code is provided as a supplement.

\subsection{Experimental Setup}

\paragraph{Datasets}
We use the IMDB$\&$Reddit Dataset  \cite{yanardag2015deep} and the Ego-network Dataset  \cite{qiu2018deepinf} in our experiments. Following the literate, we include Coreness \cite{batagelj2003m}, Pagerank \cite{page1999pagerank}, Eigenvector Centrality \cite{bonacich1987power}, Clustering Coefficient \cite{watts1998collective}, and Degree/Rarity \cite{adamic2003friends} as the node features for both datasets.

\textbf{IMDB$\&$Reddit Dataset} IMDB$\&$Reddit consists of the
IMDB-Binary and Reddit-Binary datasets, each denoting a single domain. The statistics of these datasets are described in Table \ref{tab1:plain}, and the node features are presented in Table \ref{tab2:plain}. We can see that the datasets are quite different in terms of their sizes.
\begin{itemize} 
\item  \textbf{IMDB-BINARY}  is a balanced dataset. Each graph in this dataset represents an ego network for an
actor/actress. Nodes in the graph correspond to actors/actresses.
There will be an edge between two actors/actresses who
appear in the same movie. A graph is generated from either romance   or action movies. The task is to classify the graph into romance or Action genres. A movie can belong to both categories. In this situation, we follow the precedent \cite{wang2019transferable} and remove the movie from
Romance genre while maintaining the
Action genre.
\item \textbf{REDDIT-BINARY}   is also a balanced dataset.
Each graph represents an online discussion thread. Nodes correspond to users. If one of the users has responded to
another’s comments, then an edge exists
between them. The discussion threads are drawn
from four communities: AskReddit and IAmA 
are question/answer-based communities. Atheism and TrollXChromosomes are discussion-based
communities. The binary classification task is to classify a graph into discussion-based or question/answer-based communities.
\end{itemize}
\begin{table}[!ht]
    \centering
    \small
    \parbox{.45\linewidth}{
    \begin{tabular}{lll}
        \toprule
        Dataset  &  IMDB-Binary &  Reddit-Binary \\
        \midrule
        $\#$ of Node     &  19,773     & 859,254 \\
        $\#$ of Edge     & 96,531    & 995,508  \\
        $\#$ of Graph  &  1000     & 2000 \\
        Avg Degree & 4.88     & 1.16 \\
        \bottomrule
    \end{tabular}
    \caption{Statistics of IMDB$\&$Reddit Dataset}
    \label{tab1:plain}}
    \centering
    \small
   \parbox{.45\linewidth}{ \begin{tabular}{ll}
        \toprule
        Feature Type  & Feature Name \\
        \midrule
        \multirow{6}{*}{Vertex} & Coreness. \\
             & Pagerank. \\
             & Hub and authority scores. \\
             & Eigenvector Centrality. \\
             & Clustering Coefficient. \\
             & Degree.\\
        \bottomrule
    \end{tabular}
    \caption{Node Features of IMDB$\&$Reddit Dataset.}
    \label{tab2:plain}}
\end{table}

\paragraph{Ego-network Dataset}  Ego-network consists of data from four
social network platforms, Digg, OAG, Twitter, and Weibo, each representing a    domain. Each network is modeled as a graph.
Each graph has 50 nodes, and nodes in the graphs represent users. Every graph has an ego user. An edge is drawn between two nodes if a social connection occurs between two users. The definitions of social connection of these four social network platforms are different. We extract the descriptions of social connections and social actions of each social network   according to Qui et al. \citeyear{qiu2018deepinf}:
\begin{itemize}
\item \textbf{Digg} allows
users to vote for web content such as stories and news (up or down).
The social connection is users' friendship, and the social action is voting for the content.
\item \textbf{OAG}   is generated from AMiner and Microsoft Academic Graph. The social connection is represented as the co-authorship of users, and the social action is the citation behavior.
\item \textbf{Twitter} %On Twitter, users post and interact with tweets. Tweets are messages that are no longer than 280 characters. 
currently known as X, the social connections on Twitter represent users' friendship on Twitter, and the social action is posting tweets related to the Higgs boson, a particle discovered in 2012.
\item \textbf{Weibo}  is a social platform similar to Twitter and popular among Chinese users. The Weibo dataset
 includes posting logs between September 28th, 2012, and October 29th, 2012, among 1,776,950 users. The social connection is defined as users' friendship, and the social action is re-posting messages on Weibo.
\end{itemize}

All the graphs in the four domains are labeled as active
or inactive, the ego user’s action status. If the user makes the social action, then the use is active.  The task is to identify whether the ego users are active or inactive.
The statistics of the Ego-network dataset are described in Table \ref{tab3:plain}. The features of the Ego network are presented in Table \ref{tab4:plain}.
In addition to previously referenced node features, the features also contain DeepWalk embeddings for each node \cite{perozzi2014deepwalk}, the number/ratio of active neighbors \cite{backstrom2006group}, the density of subnetwork induced by active neighbors \cite{ugander2012structural}, and the number of connected components formed by active neighbors \cite{ugander2012structural}.

\begin{table}[!ht]
\setlength{\tabcolsep}{4pt}
    \centering
    \footnotesize
 \parbox{.5\linewidth}{   \begin{tabular}{p{1.4cm}p{1.1cm}p{1.12cm}p{1.2cm}p{1.11cm}}
        \toprule
        Dataset  &  OAG & Twitter & Weibo & Digg \\
        \midrule
        $\#$ of Node & 953,675 & 456,626 & 1,776,950 &279,630\\
        $\#$ of Edge  &  4,151,463 & 12,508,413 & 308,489,739 & 1,548,126  \\
        $\#$ of Graph  &  499,848 & 499,160 & 779,164 & 244,128\\
        Avg Degree & 4.35 & 27.39 & 173.60 & 5.54 \\
        \bottomrule
    \end{tabular}
    \caption{Statistics of Ego-network Dataset}
    \label{tab3:plain}}
    \centering
    \footnotesize
   \parbox{0.44\linewidth}{ \begin{tabular}{ll}
        \toprule
        Feature Type  & Feature Name \\
        \midrule
        \multirow{6}{*}{Vertex} & Coreness. \\
             & Pagerank.  \\
             & Hub and authority scores. \\
             & Eigenvector Centrality. \\
             & Clustering Coefficient. \\
             & Rarity (reciprocal of ego user’s degree).\\
        \midrule
        Embedding & 64-dim DeepWalk embedding.\\
        \midrule
        \multirow{5}{*}{Ego-net} & The number/ratio of active neighbors.\\
        & Density of sub-network induced\\
        & by active neighbors.\\
        & Number of Connected components\\
        & formed by active neighbors.\\
        \bottomrule
    \end{tabular}
    \caption{Node Features of Ego-network Dataset}
    \label{tab4:plain}}
\end{table}

%%%%%%%%%%%%%%%%%% Results %%%%%%%%%%%%%%%%%%%

%%%%%%%%%%%%%%%%%%%% Results %%%%%%%%%%%%%%%%%

\paragraph{Baselines for Comparison}
There is a limited number of unsupervised UDA algorithms specifically designed for graph-structured data. As a result, we conduct a comparative analysis between our proposed method and updated versions of several representative methods (DANN, MDD, DIVA) and current state-of-the-art UDA methods  (SDAT, BIWAA, ToAlign) for array-structured data. To facilitate the adaptation of these algorithms to graph-structured data and consistent with the structure of our VGAE encoder, we replace the encoders originally designed for array-structured data with Graph Attention Networks (GATs). These methods are explained below:
\begin{itemize}
\item \textbf{Sources:} Plain GATs trained without domain adaptation techniques.
\item \textbf{DANN:} Domain Adversarial Neural Network (DANN) \cite{ganin2016domain} adopts an 
adversarial learning strategy. It contains a domain classifier. The domain classifier tries to distinguish the samples from which domain and
the feature extractor aims to confuse the domain classifier. A gradient reverse layer \cite{ganin2015unsupervised} is applied for optimization. 
\item \textbf{MDD:} Margin Disparity Discrepancy (MDD) \cite{zhang2019bridging} is first proposed for computer vision tasks. It measures the distribution discrepancy and is tailored
to the minimax optimization for training.
\item \textbf{DIVA:} Domain Invariant Variational Autoencoders (DIVA) \cite{ilse2020diva} disentangles the inputs into three latent variables, domain latent variables, semantic latent variables, and residual variations latent variables. It is proposed to solve problems in fields such as medical imaging. 
\item \textbf{SDAT:} Smooth Domain Adversarial Training (SDAT) \cite{rangwani2022closer} focuses on achieving smooth minima with respect to classification loss, which stabilizes adversarial training and improves the performance on the target domain.
\item \textbf{BIWAA:} Backprop Induced Feature Weighting for Adversarial Domain Adaptation with
Iterative Label Distribution Alignment (BIWAA) \cite{biwaa2023} employs a classifier-based backprop-induced weighting of the feature space, allowing the domain classifier to concentrate on features that are important for classification and coupling the classification and adversarial branch more closely.
\item \textbf{ToAlign:} Task-oriented Alignment for Unsupervised Domain Adaptation (ToAlign) \cite{wei2021toalign} decomposes features in the source domain into classification task-related and classification task-irrelevant parts under the guidance of classification meta-knowledge, ensuring that the domain adaptation is beneficial for the performance on the classification task.
\end{itemize}

% \textbf{Baselines and SOTA Methods}
% There is a limited number of unsupervised domain adaptation (UDA) algorithms specifically designed for graph-structured data. As a result, we conduct a comparative analysis between our proposed method and updated versions of several representative domain adaptation methods for array-structured data. These methods include DANN \cite{ganin2016domain}, MDD \cite{zhang2019bridging}, and DIVA \cite{ilse2020diva} and a GAT trained without any domain adaptation techniques (Source). Additionally, we consider recently published SOTA approaches such as SDAT \cite{rangwani2022closer}, BIWAA \cite{biwaa2023}, and ToAlign \cite{wei2021toalign}. To ensure a fair comparison and facilitate the adaptation of these algorithms to graph-structured data, we replace the feature extractors originally designed for array-structured data with Graph Attention Networks (GATs). 

\paragraph{Evaluation Metrics}
Following the literature, we use F1-Score  to account for imbalance in the datasets.

\paragraph{Training Scheme}
In our evaluation, we rigorously train five models for each baseline method by employing five distinct random seeds for parameter initialization and dropping or adding edges during the data augmentation phase. We report both the average performance and standard deviation of the obtained F1-score. To ensure a fair comparison across all methods, we maintain the same seed for data shuffling. The optimization process uses the Adam \cite{kingma2014adam} optimizer. Please refer to the Appendix for a comprehensive description of our training scheme.

% F1-Score is the harmonic mean of the precision and
% recall. Formally, we define the F1-score as:
% \begin{equation}
% F_1 = \frac{\text{Precision} \cdot \text{Recall}}{\text{Precision} + \text{Recall}}
% \end{equation}

% \begin{table}[ht!]
%     \centering
%     \footnotesize
%     \begin{tabular}{lll}
%         \toprule
%         Parameter  & Ego-network & IMDB$\&$Reddit \\
%         \midrule
%         Batch size & 256 & 64\\
%         Learning rate & 0.01 & 0.001\\
%         Dropout rate & 0.5 & 0.2\\
%         $\gamma$ & 0.2 & 0.2\\
%         $\lambda_1$ & 0.02 & 0.1\\
%         $\lambda_2$ & 0.02 & 0.001\\
%         $\lambda_3$ & 0.01 & 0.001 \\
%         $\lambda_4$ & 0.25 & 0.25\\
%         \bottomrule
%     \end{tabular}
%     \caption{Hyper-parameters of DNAN}
%     \label{tab9:plain}
% \end{table}

\subsection{Performance Results and Comparison} 
Tables \ref{tab5:plain} and \ref{tab6:plain} present our performance results. The bold font denotes the highest performance in each column.

\begin{table*}[ht!]
    \centering
    \small
    \begin{tabular}{p{0.75cm}p{0.85cm}p{0.75cm}p{0.75cm}p{0.75cm}p{0.75cm}p{0.75cm}p{0.75cm}p{0.75cm}p{0.75cm}p{0.75cm}p{0.75cm}p{0.75cm}p{0.5cm}}
        \toprule
        Method  & O$\rightarrow$T&  O$\rightarrow$W &   O$\rightarrow$D  & T$\rightarrow$O    & T$\rightarrow$W    & T$\rightarrow$D    & W$\rightarrow$O &  W$\rightarrow$T &    W$\rightarrow$D &    D$\rightarrow$O  & D$\rightarrow$T    & D$\rightarrow$W    & Avg\\
        \midrule
        Source & 40.0$_{\pm0.0}$ & 40.4$_{\pm0.3}$ & 43.8$_{\pm2.4}$ & 40.2$_{\pm0.0}$ & \textbf{48.0}$_{\pm1.2}$ & 41.3$_{\pm0.0}$ & 40.2$_{\pm0.0}$ & 46.6$_{\pm0.9}$ & 41.3$_{\pm0.1}$ & 40.2$_{\pm0.0}$ & 40.0$_{\pm0.0}$ & 39.8$_{\pm0.0}$ & 41.8\\
        DANN & 42.0$_{\pm0.6}$ & 41.7$_{\pm0.6}$   & 51.3$_{\pm0.8}$  & 40.7$_{\pm0.7}$ &    42.0$_{\pm0.9}$    & 49.9$_{\pm1.7}$ &   40.3$_{\pm0.1}$ &  41.3$_{\pm0.6}$    & 50.9$_{\pm0.4}$ &   40.3$_{\pm0.0}$    & 40.4$_{\pm0.3}$  & 42.5$_{\pm0.7}$  & 43.6\\
        MMD & 40.2$_{\pm0.1}$ & 41.2$_{\pm1.1}$    & 48.0$_{\pm3.2}$  & 40.2$_{\pm0.0}$  & 45.5$_{\pm1.5}$  & 41.3$_{\pm0.0}$  & 40.2$_{\pm0.0}$  & 46.2$_{\pm2.9}$  & 41.9$_{\pm1.3}$  & 40.2$_{\pm0.0}$  & 40.1$_{\pm0.1}$  & 40.0$_{\pm0.2}$ & 42.1\\
        DIVA & 42.1$_{\pm0.5}$  &42.4$_{\pm1.4}$   & 48.9$_{\pm0.7}$  & 40.3$_{\pm0.2}$  & 42.0$_{\pm0.4}$  & 50.3$_{\pm0.5}$  & 40.4$_{\pm0.3}$  & 41.2$_{\pm0.3}$  & 48.7$_{\pm0.9}$  & \textbf{40.6}$_{\pm0.5}$  & 41.3$_{\pm0.4}$  & 42.5$_{\pm0.5}$  & 43.4\\
        SDAT & 40.2$_{\pm0.1}$ & 40.1$_{\pm0.5}$ & 42.2$_{\pm1.6}$ & 40.2$_{\pm0.0}$ & 41.6$_{\pm1.3}$ & 42.9$_{\pm3.2}$ & 40.3$_{\pm0.2}$ & 41.1$_{\pm0.7}$ & 43.1$_{\pm2.8}$ & 40.2$_{\pm0.0}$ & 40.1$_{\pm0.1}$ & 39.9$_{\pm0.1}$ & 41.0\\
        BIWAA & 40.1$_{\pm0.2}$ & 41.5$_{\pm0.2}$ & 45.1$_{\pm1.3}$ & 40.3$_{\pm0.2}$ & \textbf{48.0}$_{\pm2.9}$ & 43.3$_{\pm3.0}$ & 40.2$_{\pm0.0}$ & \textbf{50.4}$_{\pm0.5}$ & 45.7$_{\pm3.8}$ & 40.3$_{\pm0.1}$ & 41.0$_{\pm0.9}$ & \textbf{43.1}$_{\pm0.3}$ & 43.2\\
        ToAlign & 36.5$_{\pm13.3}$ & 43.0$_{\pm0.9}$ & 49.1$_{\pm1.4}$ & \textbf{40.8}$_{\pm0.3}$ & 43.1$_{\pm1.1}$ & 50.2$_{\pm0.5}$ & 40.7$_{\pm0.4}$ & 42.8$_{\pm1.3}$ & 48.4$_{\pm3.2}$ & 40.5$_{\pm0.2}$ & 43.4$_{\pm0.9}$ & 42.6$_{\pm1.6}$ & 43.4\\
        \midrule
        DNAN & \textbf{42.9}$_{\pm1.6}$ & \textbf{43.4}$_{\pm1.2}$ & \textbf{53.7}$_{\pm0.6}$ & \textbf{40.8}$_{\pm0.2}$ & 45.3$_{\pm3.0}$ & \textbf{53.9}$_{\pm1.0}$ & \textbf{40.8}$_{\pm0.4}$ & 48.6$_{\pm0.8}$ & \textbf{53.4}$_{\pm2.9}$ & \textbf{40.6}$_{\pm0.2}$ & \textbf{44.1}$_{\pm1.5}$ & 42.8$_{\pm0.9}$ & \textbf{45.9}\\
        \bottomrule
    \end{tabular}
    \caption{Performance results on Ego-network Dataset}
    \label{tab5:plain}
\end{table*}

\begin{table*}[ht!]
    \centering
    \small
    \begin{tabular}{p{0.5cm}p{0.8cm}p{0.8cm}p{0.8cm}p{0.8cm}p{0.8cm}p{0.8cm}p{0.8cm}p{0.9cm}}
        \toprule
         Task  & Source & DANN  &
         MMD & DIVA & SDAT & BIWAA & ToAlign & DNAN\\
         \midrule
         I$\rightarrow$R & 63.4$_{\pm0.2}$ & 63.9$_{\pm0.8}$ & 63.7$_{\pm0.4}$ & 63.6$_{\pm0.5}$ & 63.6$_{\pm0.6}$ & 64.0$_{\pm0.8}$ & 63.3$_{\pm0.2}$ & \textbf{64.2}$_{\pm0.6}$\\
         R$\rightarrow$I & 72.3$_{\pm1.7}$ & 72.0$_{\pm1.7}$ & 73.6$_{\pm1.7}$ & 71.1$_{\pm0.3}$ & 74.1$_{\pm2.0}$ & 71.4$_{\pm1.0}$ & 73.4$_{\pm0.8}$ & \textbf{74.9}$_{\pm2.0}$\\
         \midrule
         Avg & 67.8 & 68.0 & 67.3  & 68.0 & 68.8 & 67.7 & 68.3 & \textbf{69.6}\\
        \bottomrule 
    \end{tabular}
    \caption{Perofmrance results on IMDB$\&$Reddit Dataset}
    \label{tab6:plain}
\end{table*}

\paragraph{Ego-network Results} Results for this dataset are presented in  Table \ref{tab5:plain}. 
In this benchmark, twelve UDA tasks can be defined by pairing the four domains. Our experimental results indicate that the DNAN   performs the best on average and achieves state-of-the-art performance on nine tasks: O to T, O to W, O to D, T to O, T to D, W to O, W to D, D to O, and D to T. DNAN has good performances on T to W and W to T, and achieve SOTA performance on D to T and T to D tasks, showing that DNAN can successfully handle similar domains, as Digg, Twitter, and Weibo are similar content-sharing platforms. Notably, it exceeds the second-best methods by about $4\%$ on T to D and about $3\%$ on W to D. In addition, DNAN can also achieve SOTA performance when there is a large distribution gap between domains, such as on tasks between OAG and Twitter or OAG and Weibo. It is important to underscore that no single method can achieve the best performance on all tasks, likely due to the diverse range of domain gaps. %Except for O to T and W to D, our approach is still competitive on the tasks where our approach does not lead to the best results.

\paragraph{IMDB$\&$Reddit Results} %The experiment results on the IMDB$\&$Reddit dataset are presented in Table 8. 
Results for this dataset are presented in Table \ref{tab6:plain}. We note that the experiments demonstrate that the UDA methods perform better on the Reddit to IMDB task (R to I) than on the IMDB to Reddit task (I to R), indicating that the two tasks are not equally challenging. We hypothesize that the smaller size of   IMDB-Binary compared to   Reddit-Binary may result in performance degradation when testing on the larger Reddit-Binary dataset, as less knowledge can be transferred to the target domain. Our experiment results show that   DNAN   outperforms all other methods on both the R to I and I to R tasks, leading to SOTA results on average. On the I to R task, we observe that all methods perform similarly. Though DNAN does not outperform other methods by a large margin on the I to R task, the results still indicate that DNAN has a competitive performance compared to other methods. 
It is worth noting that the performance of a UDA algorithm may vary to some extent based on hyperparameter tuning. Therefore, when comparing two UDA algorithms with similar performance, they should be considered equally competitive. Based on this consideration, we can conclude that our proposed method performs competitively on all the UDA tasks and outperforms other UDA methods on average. These findings suggest that DNAN can serve as a robust UDA algorithm.% solution for various domain adaptation tasks.

\subsection{Analytic and Ablative and  Experiments}
We first perform analytic experiments to offer a deeper insight into our approach. We then performed an ablative experiment to demonstrate that all components in our algorithm are important to achieve optimal performance.

\begin{figure*}
\centering
\includegraphics[width=17cm]{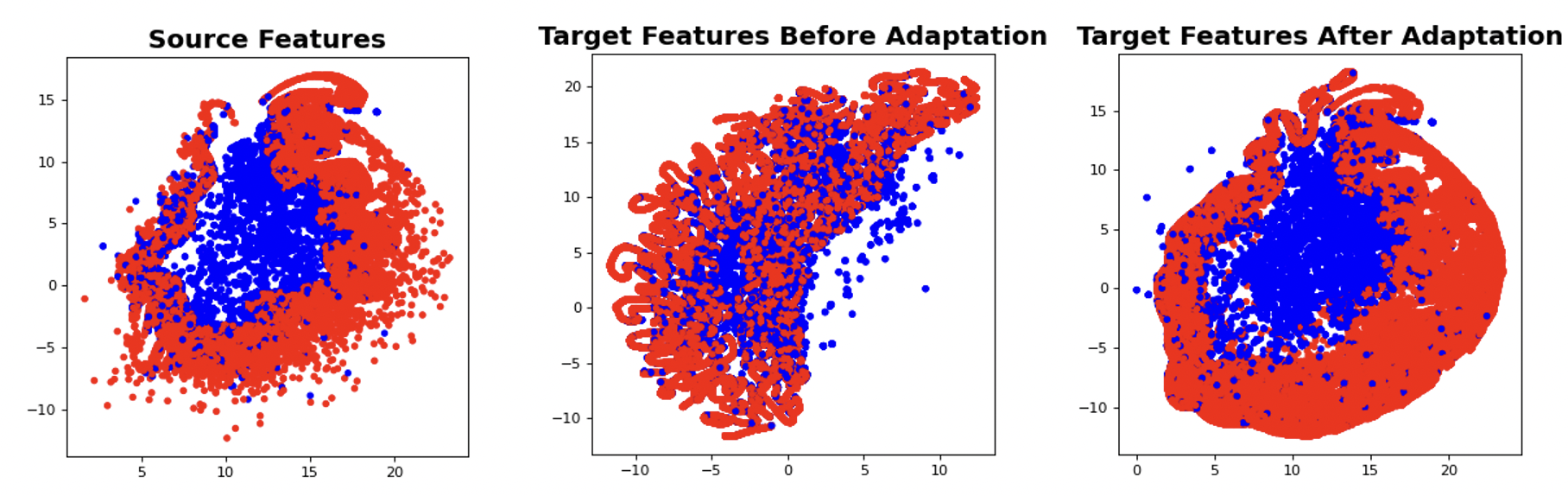}
\caption{UMAP visualization for the representation of the testing split of data on the Oag to Weibo task
. The blue and red points denote each of the classes. The left and the center plots visualize the features of the source and the target domains extracted by the source-trained model before domain adaptation. The right plot visualizes the features of the target domain after adaptation is performed using DNAN. } 
\label{fig2}
\end{figure*}

\paragraph{The effect of DNAN on data representations in the embedding space} 
To assess the effectiveness of our proposed approach,
we investigated the impact of DNAN on the distribution of the target domain in the shared embedding space on the Oag to Weibo task (O to Weibo). We picked this task to demonstrate the effect of our model because it is challenging as Weibo and Oag are very dissimilar platforms; one is connected by co-authorship, and the other is friendship. We utilized the UMAP~\cite{mcinnes2018umap} visualization tool to reduce the dimensionality of the data representations to two for 2D visualization. We then visualized the testing splits of the source domain data, the target domain data before adaptation, and the target domain data after adaptation using DNAN. In Figure~\ref{fig2}, each point represents a single data point in the output space of the feature extractor subnetwork, and blue/red colors denote the two classes. The figures reveal that the distribution of the target domain is not well aligned with the distribution of the source domain before adaptation, and the class boundary is unclear as most of the red dots are mixed with the blue dots. This discrepancy translates into performance degradation. However, the distribution of the target domain aligns much better with the source domain after domain adaptation using DNAN. This visualization demonstrates that DNAN successfully aligns the two distributions in the embedding space, mitigating the effects of domain shift.

\begin{table*}[ht!]
    \centering
    \small
    \begin{tabular}{{p{1.4cm}p{0.73cm}p{0.73cm}p{0.73cm}p{0.73cm}p{0.73cm}p{0.73cm}p{0.73cm}p{0.73cm}p{0.73cm}p{0.73cm}p{0.73cm}p{0.73cm}p{0.4cm}}}
        \toprule
        Method  & O$\rightarrow$T&  O$\rightarrow$W &   O$\rightarrow$D  & T$\rightarrow$O    & T$\rightarrow$W    & T$\rightarrow$D    & W$\rightarrow$O &  W$\rightarrow$T &    W$\rightarrow$D &    D$\rightarrow$O  & D$\rightarrow$T    & D$\rightarrow$W    & Avg\\
        \midrule
        DNAN-D & 42.8$_{\pm1.3}$ & 42.5$_{\pm1.7}$ & 52.3$_{\pm2.6}$ & 40.5$_{\pm0.2}$ & \textbf{46.9}$_{\pm2.0}$ & 50.0$_{\pm3.5}$ & 40.4$_{\pm0.2}$ & \textbf{50.1}$_{\pm0.6}$ & 52.8$_{\pm2.4}$ & \textbf{40.6}$_{\pm0.2}$ & 42.1$_{\pm1.3}$ & 42.5$_{\pm1.8}$ & 45.4\\
        DNAN-N & \textbf{44.4}$_{\pm2.2}$ & 43.1$_{\pm0.7}$ & 52.6$_{\pm1.0}$ & \textbf{41.0}$_{\pm0.3}$ & 45.0$_{\pm2.4}$ & 52.7$_{\pm2.7}$ & 40.7$_{\pm0.3}$ & 46.9$_{\pm2.2}$ & 53.3$_{\pm2.5}$ & 40.5$_{\pm0.2}$ & 43.0$_{\pm1.1}$ & \textbf{43.6}$_{\pm0.9}$ & 45.6\\
        \midrule
        DNAN &  42.9$_{\pm1.6}$ & \textbf{43.4}$_{\pm1.2}$ & \textbf{53.7}$_{\pm0.6}$ & 40.8$_{\pm0.2}$ & 45.3$_{\pm3.0}$ & \textbf{53.9}$_{\pm1.0}$ & \textbf{40.8}$_{\pm0.4}$ & 48.6$_{\pm0.8}$ & \textbf{53.4}$_{\pm2.9}$ & \textbf{40.6}$_{\pm0.2}$ & \textbf{44.1}$_{\pm1.5}$ & 42.8$_{\pm0.9}$ & \textbf{45.9}\\
        \bottomrule
    \end{tabular}
    \caption{Ablation Study Results on Ego-network Dataset}
    \label{tab7:plain}
\end{table*}

\begin{table}[ht!]
\setlength{\tabcolsep}{3pt}
    \centering
    \small
    \begin{tabular}{llll}
        \toprule
        Task  & DNAN-D & DNAN-N & DNAN\\
        \midrule
        I to R & 63.8$_{\pm0.4}$ & 64.0$_{\pm0.5}$ & \textbf{64.2}$_{\pm0.6}$\\
        R to I & 72.3$_{\pm2.4}$ & 74.2$_{\pm1.8}$ & \textbf{74.9}$_{\pm2.0}$\\
        \midrule
        Avg &  68.0 & 69.0 &\textbf{69.6}\\
        \bottomrule
    \end{tabular}
    \caption{Ablation Study Results on IMDB$\&$Reddit Dataset}
    \label{tab8:plain}
\end{table}

\textbf{Ablative study} The ablation experiments are conducted to demonstrate the effectiveness of the two ideas we benefit from to develop DNAN. To this end, we remove one of the two components at a time and report our performance. We denote the ablated versions of DNAN as:
 (i) \textbf{DNAN-D} We exclude the denoising mechanism and only apply NWD loss. 
and (ii) \textbf{DNAN-N} We exclude NWD loss and only apply the denoising mechanism.

  Our ablation study results for the Ego-network and the IMDB$\&$Reddit  datasets are illustrated in Table \ref{tab7:plain} and \ref{tab8:plain}, respectively. The Ego-network dataset results reveal that the integration of both NWD loss and the denoising mechanism (DNAN) yields the highest average performance at 45.9$\%$. The DNAN-D configuration, lacking the denoising mechanism, shows competitive performance with an average of 45.4$\%$. However, the DNAN-N configuration, which excludes NWD loss, displays an even smaller decrease in performance, with an average of 45.6$\%$. For the IMDB$\&$Reddit dataset, the full DNAN model again demonstrates superior performance with an average score of 69.6$\%$. Interestingly, the DNAN-N variant outperforms DNAN-D with averages of 69.0$\%$ and 68.0$\%$, respectively. This observation indicates that the denoising mechanism is more critical in this context. 
  The ablation study highlights the importance of both the denoising mechanism and NWD loss in our proposed method. While the NWD loss and the denoising techniques contribute more evenly to the Ego-network dataset, the denoising mechanism is more beneficial for the IMDB$\&$Reddit dataset. This suggests that the effectiveness of each component is context-dependent, and future work may explore this dependency in greater depth. The cooperating effect of combining both techniques confirms the robustness of our DNAN model, as it consistently outperforms its counterparts with either component excluded.
 
%  Comparing the performance results of DNAN-W and DNAN-M with those of the original DNAN method, we observe that the combination of SWD and MMD leads to the best performance on both datasets. Although removing either SWD or MMD results in reduced performance, removing SWD leads to more significant performance degradation than removing MMD. This observation is expected since SWD aligns both the high-order and low-order statistical moments, while MMD only accounts for low-order moments. Therefore, combining these two techniques provides a more comprehensive approach to domain alignment, leading to superior performance on both datasets. 

% By inspecting the results for DNAN-D, we observe that the effectiveness of data augmentation varies between the two datasets. It improves the average performance on the IMDB$\&$Reddit dataset but decreases the average performance on the Ego Network dataset. However, its negative impact is marginal, even when not helpful, since DNAN remains competitive, as evidenced by the results presented in Table ~\ref{tab5:plain}. This finding highlights that data argumentation is beneficial on average.

\begin{figure*}
\centering
\includegraphics[width=16cm]{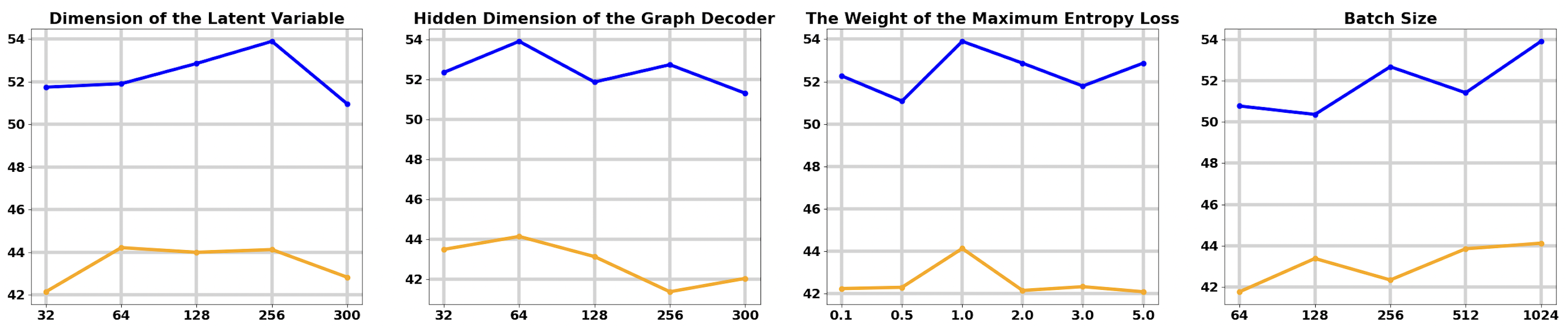}
\caption{The performance of DNAN with different hyperparameter settings on Twitter to Digg (Blue lines) and Digg to Twitter (Yellow lines) tasks.} 
\label{fig3}
\end{figure*}

\subsection{Hyperparameters Sensitivity Analysis}
An important concern for most algorithms is tuning the hyperparameters and measuring the performance sensitivity with respect to them. We evaluate the sensitivity of   DNAN   with respect to various hyperparameters on two tasks: Twitter to Digg (T to  D) and Digg to Twitter (D to  T). We varied the dimension of the latent embedding space, the hidden dimension of the graph decoder, the weight of the maximum entropy loss, and the batch size. We present the F1-scores of the DNAN model as a linear function of these hyperparameters in Figure \ref{fig3}, with the blue lines representing the T to  D task and the yellow lines representing D to  T task. Through inspecting this figure, we deduce:
\begin{itemize}
\item \textbf{Dimension of the Latent Embedding Space:}
We test the performance of DNAN on five different dimension sizes for the embedding space: 32, 64, 128, 256, and 300. The performance of DNAN peaks at a latent variable size of 256 for the T to  D task and a less pronounced peak on the D to T task, indicating that a moderately large value for the dimension of the latent variable is beneficial for capturing the salient features of the data. Performance declines when the dimension is too small to capture the complexity or too large, potentially introducing noise or overfitting. We note, however, that the result indicates that the performance remains relatively decent for a wide range of embedding sizes.
\item \textbf{Hidden Dimension of the Graph Decoder:}
Similar to the experiments on the dimension of the latent variable, we test the performance of DNAN on five dimension sizes: 32, 64, 128, 256, and 300. The hidden dimension of the graph decoder shows a performance peak at 64 for the T to D task and the D to  T task. This observation suggests that a moderately small representation capacity in the graph decoder is more beneficial. Compared with performances on the D to T task, the T to D task is less sensitive to this hyperparameter.
\item \textbf{Weight of the Maximum Entropy Loss:}
We test the DNAN on six weights: 0.1,0.5,1.0,2.0,3.0,5.0. The weight of the maximum entropy loss presents a clear peak at 1.0 for both the T to  D and D to T tasks, suggesting that a balanced contribution of the entropy loss is critical for performance.
\item \textbf{Batch Size:} We test five batch sizes: 64, 128, 256, 512, 1024.
For batch size, there is a trend of increasing performance as the size grows, with a notable peak at a batch size of 1024 for the T to  D and D to T tasks. This implies that the performance of DNAN  benefits from larger batch sizes, possibly due to more stable gradient estimates. Compared with the T to D task, the D to T task is less affected by batch size variations.
\end{itemize}

The sensitivity analysis of hyperparameters for the DNAN model on the T to D and D to T tasks demonstrates the stability of DNAN models when using different hyperparameter values, as there is only a moderate fluctuation around $\pm 3 \%$. However, fine-tuning hyperparameters to the specific characteristics of the task and the dataset is beneficial. Although optimal performance is achieved with a latent variable of 256, a hidden decoder dimension of 64, an entropy loss weight of 1.0, and a batch size of 1024, tuning the hyperparameter is not necessary to achieve a good performance.

\begin{table}[!ht]
\setlength{\tabcolsep}{4pt}
    \centering
    \footnotesize
 \parbox{.5\linewidth}{ \begin{tabular}{p{0.5cm}p{0.6cm}p{0.6cm}p{0.6cm}p{0.6cm}p{0.8cm}p{0.8cm}p{0.6cm}}
        \toprule
         Task & DANN  &
         MMD & DIVA & SDAT & BIWAA & ToAlign & DNAN\\
         \midrule
         I$\rightarrow$R & 10 & 260 & 14 & 12 & 899 & 10 & 10  \\
         R$\rightarrow$I & 3 & 2 & 5 & 6 & 34 & 3 &  5\\
        \bottomrule 
    \end{tabular}
    \caption{Training time for IMDB$\&$Reddit Dataset}
    \label{tab9:plain}}
    \centering
    \footnotesize
   \parbox{0.44\linewidth}{   \begin{tabular}{p{1.6cm}p{2.5cm}}
        \toprule
         Methods & Model Size\\ 
         \midrule
         DANN  & (K+3M+3)M\\
         MMD & (K+4M+4)M \\
         DIVA & (K+11M+3D+3)M \\
         SDAT & (K+3M+4)M\\
         BIWAA & (K+3M+3)M\\
         ToAlign & (K+2M+11)M\\
         DNAN & (K+4M+D+2)M\\
        \bottomrule 
    \end{tabular}
    \caption{Model complexity on IMDB$\&$Reddit-Binary dataset. K represents the input features dimension, M represents the hidden dimension, and D represents the output dimension of the Decoder.}
    \label{tab10:plain}}
\end{table}

\subsection{Time Complexity and Model Complexity Analysis}
The analysis of the training time and model complexity for various domain adaptation methods on the IMDb$\&$Reddit dataset is presented in Table ~\ref{tab9:plain} and Table ~\ref{tab10:plain}.

\begin{itemize}
\item \textbf{Training Time:}
The training times reported in Table ~\ref{tab9:plain} illustrate the efficiency of the DNAN model relative to its counterparts. For the I  to  R task, DNAN required 10 minutes, positioning it the fastest in terms of training time, together with DANN and ToAlign, compared to other methods. In the R  to  I task, DNAN again demonstrated moderate efficiency with 5 minutes, with MMD being the fastest at 2 minutes and BIWAA the slowest at 34 minutes. These results suggest that DNAN provides a balanced trade-off between model performance and training efficiency without adding a significant computational overload.
\item \textbf{Model Complexity:}
The model complexity, as shown in Table ~\ref{tab10:plain}, is assessed based on the number of parameters in the models, which is a function of the input features dimension (K), hidden dimension (M), and the output dimension of the Decoder (D). Compared to other methods like DANN and SDAT, which have similar forms, DNAN introduces additional complexity due to the parameters in the graph decoder. However, it remains less complex than DIVA, which includes an extra (7M+2D+1)M term.
\end{itemize}
We conclude that the DNAN model shows competitive training time that is significantly lower than the most time-consuming method (BIWAA) while maintaining comparable or better performance. Model complexity analysis reveals that DNAN, while not the simplest, avoids the higher complexity seen in more complex methods such as DIVA, balancing the computational cost with the capacity to learn and transfer knowledge effectively for better UDA performance. This observation is important because, in some applications, UDA should be performed fast because the input distribution changes relatively continually, and the time needed to update the model is limited.

% \begin{table*}[ht!]
%     \centering
%     \small
%     \begin{tabular}{p{0.5cm}p{0.6cm}p{0.6cm}p{0.6cm}p{0.6cm}p{0.8cm}p{0.8cm}p{0.6cm}}
%         \toprule
%          Task & DANN  &
%          MMD & DIVA & SDAT & BIWAA & ToAlign & DNAN\\
%          \midrule
%          I$\rightarrow$R & 10 & 260 & 14 & 12 & 899 & 10 & 7  \\
%          R$\rightarrow$I & 3 & 2 & 5 & 6 & 34 & 3 &  8\\
%         \bottomrule 
%     \end{tabular}
%     \caption{Training time for IMDB$\&$Reddit Dataset}
%     \label{tab9:plain}
% \end{table*}

% \begin{table*}[ht!]
%     \centering
%     \small
%     \begin{tabular}{p{1.6cm}p{2.5cm}}
%         \toprule
%          Methods & Model Size\\ 
%          \midrule
%          DANN  & (K+3M+9)M\\
%          MMD & (K+4M+10)M \\
%          DIVA & (K+12M+18+3D)M \\
%          SDAT & (K+3M+9)M\\
%          BIWAA & (K+4M+8)M\\
%          ToAlign & (K+2M+17)M\\
%          DNAN & (K+12M+18+3D)M\\
%         \bottomrule 
%     \end{tabular}
%     \caption{Model complexity for all methods. K  represents the input features dimension, m represents the hidden dimension of the GAT, and D represents the hidden dimension of the Decoder.}
%     \label{tab10:plain}
% \end{table*}

\section{Conclusions}
We developed a new UDA method, which is specifically designed for graph-structured data. Our two novel ideas include denoising and using the Nuclear-Norm Wasserstein Adaptation Network (DNAN) for domain alignment in a shared embedding space. Our experiments demonstrate our approach to be a promising method. By innovatively combining domain alignment through Nuclear-norm Wasserstein discrepancy with a denoising mechanism via a variational graph autoencoder, DNAN has outperformed state-of-the-art methods across two major benchmarks without adding significant computational overload.  The ability of our method to handle subtle and significant domain differences showcases its versatility and robustness. From ablative studies, the two ideas that DNAN benefits from are proven to be crucial for optimal performance.  Future work can explore extending our approach to partial domain adaptation scenarios or situations where the source domain data can not be directly accessible.
\bibliography{main}
\bibliographystyle{tmlr}

\newpage
\appendix

\section{Nuclear-norm Wasserstein Discrepancy}
\paragraph{From Intra-class and Inter-class Correlations to Domain Discrepancy} Consider a prediction matrix $P \in \mathbb{R}^{b\times k}$ predicted by classifier $C$, where $b$ represents the number of samples and $k$ represent the number of classes. $P$ has the following properties:
\begin{equation}
\sum_{j=1}^k P_{ij} = 1,  \text{  } P_{ij} \geq 0, \text{   } \forall i \in 1,2,...b
\end{equation}
The self-correlation matrix $R \in \mathbb{R}^{k\times k}$ then can be computed by $R = Z^TZ$. The intra-class correlation $I_a$ is defined as the sum of the main diagonal elements in $R$, and the inter-class correlation $I_e$ is defined as the sum of the off-diagonal elements in  $R$:
\begin{equation}
I_a = \sum_{i,j=1}^k R_{ij}, \text{ } I_e = \sum_{i\neq j}^k R_{ij}
\end{equation}
The $I_a$ and $I_e$ are very different for source and target domains. For the source domain, the $I_a$ is large while the $I_e$ is relatively small, as we train with labels available so that most samples are correctly classified. For the target domain, the $I_a$ is small while the $I_e$ is relatively large due to the lack of supervised training. Based on linear algebra, we can represent $I_a = \|P\|_F$, the Frobenius norm of $P$, and
\begin{equation}
I_a - I_e = 2\|P\|_F - b
\end{equation}
For the source domain, $I_a - I_e$ will be large; for the target domain, $I_a - I_e$ will be small. Therefore, $I_a - I_e$ can represent the discrepancy between two domains. Since the prediction matrix $P$ is generated by the classifier $C$, we can rewrite $P = C(Z)$, where $Z$ is the feature representation of the sample from either the source or the target domain. With inspiration from WGAN \cite{arjovsky2017wasserstein} and 1-Wasserstein distance, the domain discrepancy can be formally formulated as
\begin{equation}
W_F(D_s,D_t) = \sup_{\|\|C\|_F\|_L \leq K} \mathbb{E}_{Z_s \sim D_s}[\|C(Z_s)\|_F] - \mathbb{E}_{Z_t \sim D_t}[\|C(Z_t)\|_F]
\end{equation}
We call $W_F(D_s,D_t)$ the Frobenius norm-based 1-Wasserstein distance, where $D_s$ denotes the source domain, $D_t$ denotes the target domain, $\|\cdot\|_L$ denotes the Lipschitz semi-norm \cite{villani2009optimal}, and $K$ denotes the Lipschitz constant. 

\paragraph{From Frobenius Norm to Nuclear Norm} From the domain discrepancy formulated above, we can see that the classifier $C$ works like a discriminator in GAN. Therefore, we can perform adversarial training to train the feature generator via $W_F(D_s,D_t)$. However, adversarial training with $W_F(D_s,D_t)$ limits the diversity of predictions. This is because it tends to push the samples in a class with fewer samples near the decision boundary closer to a neighboring class with a significantly larger number of samples far from the decision boundary \cite{cui2021fast}. To address this limitation, the author proposes to use the nuclear norm instead of the Frobenius norm. The nuclear norm has been shown to be bound by the Frobenius norm \cite{chen2022reusing}. In addition, maximizing the nuclear norm maximizes the rank of the prediction matrix $P$ when $\|\cdot \|_F$ is nearby $\sqrt{b}$ \cite{cui2020towards,cui2021fast}. In consequence, the diversity of predictions will be enhanced. Thus, the domain discrepancy can be improved to be 
\begin{equation}
W_N(D_s,D_t) = \sup_{\|\|C\|_*\|_L \leq K} \mathbb{E}_{Z_s \sim D_s}[\|C(Z_s)\|_*] - \mathbb{E}_{Z_t \sim D_t}[\|C(Z_t)\|_*]
\end{equation}
$W_N(D_s, D_t)$ is called the Nuclear-norm 1-Wasserstein discrepancy (NWD). To integrate NWD into implementation, we can approximate the empirical NWD $\bar{W}_N$ by maximizing $\mathcal{L}_{\text{nwd}}$ that is defined below
\begin{equation}
\mathcal{L}_{\text{nwd}} = \frac{1}{N_s}\sum_{k=1}^{N_s} \|C(Z^k_s)\|_* - \frac{1}{N_t}\sum_{k=1}^{N_t}\|C(Z^k_t)\|_*, \text{  }  \bar{W}_N(D_s,D_t) \approx  \max \mathcal{L}_{\text{nwd}}
\end{equation}
where $Z^k_s$ is the feature representation of the $k$th sample in the source domain dataset and $Z^k_t$ is the feature representation of the $k$th sample in the target domain dataset. $N_s$ and $N_t$ represent the number of samples in the source and target domain, respectively.

\section{Implementation Details of DNAN}
In this section, we present our implementations of DNAN. Our codes are in Python, mainly with PyTorch ~\cite{NEURIPS2019_9015} and PyTorch Geometric ~\cite{Fey/Lenssen/2019} libraries. We train five models for every baseline using five random seeds for parameter initialization. The five random seeds are 27, 28, 29, 30, and 31. We also conducted a hyperparameter search, described in the hyperparameter sensitivity section in the paper, to find suitable hyperparameters for optimal performance. The hyperparameters we use to achieve the results listed in the main paper are presented in Table \ref{tab11:plain}.
\begin{table}[!ht]
    \centering
    \small
    \begin{tabular}{lll}
        \toprule
        Parameter  & Ego-network & IMDB$\&$Reddit \\
        \midrule
        Batch size & 1024 & 64\\
        Learning rate & 0.01 & 0.001\\
        Dropout rate & 0.5 & 0.2\\
        Encoder hidden size & 256 & 128 \\
        Decoder hidden size & 64 & 128 \\
        Learning decay rate & 0.75 & 0.75\\
        Entropy weight & 1.0 & 1.0\\
        Weight decay & 0.0005 & 0.0005 \\
        $p_\text{add}$ & 0.1 & 0.1\\
         $p_\text{drop}$ & 0.1 & 0.1\\
        \bottomrule
    \end{tabular}
    \caption{Hyper-parameters of DNAN}
    \label{tab11:plain}
\end{table}

\end{document}